\documentclass{bmvc2k}
\usepackage{multirow}
\usepackage{xspace}
\usepackage{amssymb}
\usepackage{pifont}
\usepackage{hyperref}
\usepackage{atbegshi}
\newcommand{\cmark}{\ding{51}}%
\newcommand{\xmark}{\ding{55}}%
\newcommand*{\eg}{e.g.\@\xspace}

\renewcommand{\paragraph}[1]{\vspace{0.1cm}\noindent\textbf{#1}\quad}

\makeatletter
\newcommand{\printfnsymbol}[1]{%
  \textsuperscript{\@fnsymbol{#1}}%
}
\makeatother

\title{ALBA: Reinforcement Learning for Video Object Segmentation}

\addauthor{Shreyank N Gowda*} {S.Narayana-Gowda@sms.ed.ac.uk}{1}
\addauthor{Panagiotis Eustratiadis*}{P.Eustratiadis@sms.ed.ac.uk}{1}
\addauthor{Timothy Hospedales}{t.hospedales@ed.ac.uk}{1}
\addauthor{Laura Sevilla-Lara}{l.sevilla@ed.ac.uk}{1}
\addinstitution{School of Informatics, \\University of Edinburgh, UK\\}
\runninghead{GOWDA et al.}{ALBA: Reinforcement Learning for VOS}

\def\eg{\emph{e.g}\bmvaOneDot}

\newcommand{\alg}{ALBA }
\newcommand{\sota}{state-of-the-art }

\newcommand{\specialcell}[2][c]{%
  \begin{tabular}[#1]{@{}c@{}}#2\end{tabular}}
\begin{document}

\maketitle

\begin{abstract}
We consider the challenging problem of zero-shot video object segmentation (VOS). That is, segmenting and tracking multiple moving objects within a video fully automatically, without any manual initialization. We treat this as a grouping problem by exploiting object proposals and making a joint inference about grouping over both space and time. We propose a network architecture for tractably performing proposal selection and joint grouping. Crucially, we then show how to train this network with reinforcement learning so that it learns to perform the optimal non-myopic sequence of grouping decisions to segment the whole video. Unlike standard supervised techniques, this also enables us to directly optimize for the non-differentiable overlap-based metrics used to evaluate VOS. We show that the proposed method, which we call \alg outperforms the previous \sota on three benchmarks: DAVIS 2017~\cite{davis2017}, FBMS~\cite{fbms} and Youtube-VOS~\cite{vis}.
\end{abstract}

% \vspace{0.2cm}
\begin{figure}[h]
     \includegraphics[width=\textwidth]{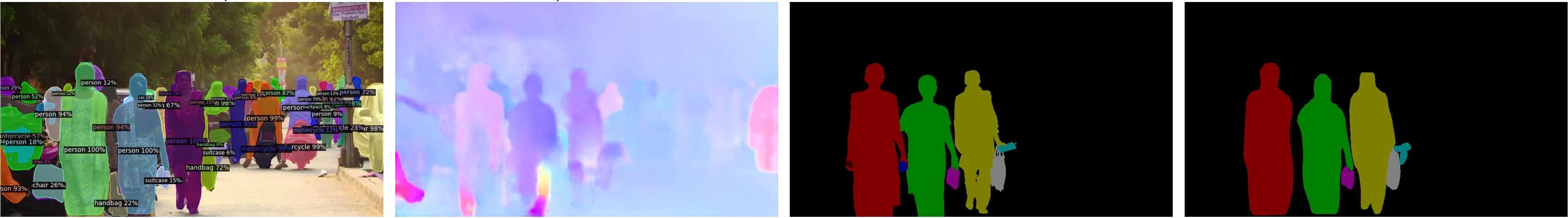}
     \caption{The proposed \alg for zero-shot video segmentation takes as input the region proposals (left), optical flow (middle left) selects and groups over time the moving ones (middle right), to mimic the ground truth (right).}
\end{figure}

\section{Introduction}
\label{sec:intro}
%Object segmentation in video is an important and long-studied problem that has been considered in many settings (zero-shot \cite{zeng2019dmm}, manually initialized \cite{xu2019mhp}), and with many types of approaches -- including fully convolutional networks that make pixel-wise decisions \cite{} and proposal grouping networks \cite{xu2019mhp,zeng2019dmm}.
Video object segmentation (VOS) aims to segment and track %multiple moving 
objects within a video. This can be thought of as a grouping problem, where a video is a collection of image regions (pixels, parts or objects) and segmentation is the selection and assignment or grouping of these regions together within and across frames. 

Region selection can be informed by appearance cues like, ``this object is likely to be moving, because it is a dog"; and temporal features learned from the optical flow could learn things like ``this object is moving independently". 
At the same time, region grouping may also be informed by appearance cues, for example assigning adjacent or similar-looking regions to the same object or leveraging knowledge like  person-like regions are more likely to appear \emph{above} bicycle-like regions than vice-versa. Grouping may also benefit from temporal information through alignment of the current frame with the previous frame, as offset by optical flow. 
Taking such a grouping perspective, we would ideally like to find an assignment of labels to regions that makes a coherent decision about region grouping both spatially within frames and temporally, across frames. However, exhaustively considering the entire joint space of potential assignments to regions is intractable, even for videos just a few seconds long.  

In the deep learning era, the majority of studies~\cite{fseg} address the VOS problem at the pixel level, with fully convolutional networks that perform dense pixel-wise classification at every frame. Spatial coherence is maintained through convolution or recurrence \cite{RVOS,fseg}, and temporal coherence is maintained through recurrence \cite{RVOS}, memory \cite{LVO} or attention \cite{MAT}. All these methods address the grouping aspect of the VOS problem in an implicit way.A minority of studies have considered explicit grouping based on object proposals, focusing on issues such as: learning good representations to inform the grouping decision (\eg from flow, images, saliency, etc) 
%\ls{the RVOS~\cite{RVOS} also start from object proposals, do you want to move them here?} 
\cite{zeng2019dmm,li2018video}, and efficiently searching the joint space of groupings to find solutions that approximate the global optimum such as via multi-hypothesis tracking \cite{xu2019mhp} or differentiable analogies to the Hungarian algorithm \cite{zeng2019dmm}.

Despite extensive excellent work in both families of approaches, existing methods have limitations that arise from the near ubiquitous use of supervised learning for training. In particular: (1) For models that sequentially consider individual frames or grouping decisions, training with supervised learning provides a kind of teacher-forcing \cite{renzato2016sequenceTrainingRNN,bengio2015scheduled}. That is, from the perspective of any given decision, the preceding decisions benefit from supervision during training in a way that they do not during testing. This creates a mismatch between training and testing conditions termed exposure bias \cite{renzato2016sequenceTrainingRNN}. This results in increased error at testing, since the model has not been trained to make optimal decisions when using its own open-loop predictions as history. (2) The use of supervised learning means that prior work is almost always trained by pixel-wise cross-entropy loss. This provides a strong form of supervision that is easy to differentiate from end-to-end learning. However optimising pixel-wise loss does not necessarily maximise the segment-overlap type of metrics of interest in VOS. 

To address these issues, we introduce a reinforcement learning-based method for VOS. Training with reinforcement learning (RL) enables us to (1) Train the model to directly optimize the intersection over union (IoU)-style metrics of interest in VOS, (2) Optimize for a policy that makes the best sequence of \emph{hard} assignment decisions both within and across-frames. Crucially, RL training enables these decisions to be non-myopic, rather than greedy \cite{zeng2019dmm}. A non-myopic grouper makes each decision to maximise the expected overall segmentation performance of a video, rather than merely the expected result of an individual grouping decision.

Specifically, we propose a simple and effective framework for segmentation by proposal grouping. Our simple relational  \cite{santoro2017relationNet,sung2018relationNet} architecture takes as input common cues (image, flow) along with the current proposal and the groups so far. It then computes the relation (group or not) between each proposal and available groups. This recurrent assignment of labels to proposals enables efficient joint decision-making about the within-frame grouping -- exploiting shape, appearance, and motion cues. Temporally-coherent assignment is further enabled through a very light-weight temporal recurrence.

%First, these networks are trained through traditional supervised learning, therefore using a differentiable function. However, the most common metric in segmentation is the intersection over union or ``IoU", which is not differentiable, and as a result, these methods need to find approximations. \ls{check this too}
%Second, the decision space of the grouping problem (\ie, deciding whether two regions are part of the same object or not is discreet. \ls{expand here}
%Third, during training, \ls{correct mistakes?} 

%In this paper, we propose to use reinforcement learning (RL) instead of traditional supervised learning. This allows us to 1) optimize the IoU metric directly and 2) learn from discreet choices. We use a fairly simple network, which takes in as input common cues (\eg image, flow and object proposals) and addresses the problem of improving the optimization of the joint assignment of object proposals using RL. 

We call the proposed method \alg (for {\em Advanced Learning for Boosted Accuracy}). Without bells and whistles, we observe that RL with  \alg  improves 6\% over using traditional supervised learning on DAVIS 2017. The final results outperform the \sota in the DAVIS 2017~\cite{davis2017} validation set, FBMS~\cite{fbms} and the recent large-scale Youtube-VOS~\cite{vis}.

\section{Related Work}
%\tim{Hopefully Laura can write this section. I don't know enough to write it easily. My notes-to-self in literature summary.tex may or may not help you.}

%\paragraph{Video Object Segmentation}
The video object segmentation problem is studied in different flavours and under different constraints. Depending on the type of labels at training time, we find instance segmentation, semantic segmentation, plenoptic segmentation or motion segmentation. Depending on the input at test time, we find one-shot segmentation (where the initial frame ground truth is given), interactive, or zero-shot. Depending on the type of scenes, we find single-object and multi-object segmentation. While ideas may be applicable to more than one of these problem categories, for simplicity we focus on the most relevant to us, which are those that address the same problem definition as us: zero-shot, multiple-object, instance segmentation. %%Since one of our main contributions is the use of reinforcement learning, we also give a brief overview of how it has been used in other related problems in computer vision. 

\paragraph{Zero-Shot Object Segmentation in Video. }
The zero-shot (also referred to as ``unsupervised") multi-object setting in video segmentation provides specific challenges. The first is the selection of the right objects to track. While in one-shot (or ``semi-supervised") the relevant objects are given, in the zero-shot they need to be discovered. Research efforts have taken a variety of approaches to address this issue, including the use of annotated eye-tracking data~\cite{AGS} that contains information about the importance of objects, modelling saliency~\cite{PDB} in the appearance, or learning attention~\cite{MAT}. The second is the tracking of multiple instances of objects, where object identities are to be matched over time. While the single-object segmentation problem can be solved with little or no temporal information~\cite{osvos}, matching objects across frames requires more sophisticated temporal modelling. Recent efforts have leveraged recurrent networks~\cite{RVOS, PDB}, graphs~\cite{AGNN} or attention~\cite{andiff}. While these efforts have yielded great progress in the benchmarks over the years, we may say that most of the focus has been on enriching representations through elaborate models, keeping the optimization process fixed, as supervised learning. Instead, our approach is to use a fairly simple representation, and focus on the optimization.

% AGS \cite{AGS}: uses additional annotated eye-tracking data to model attention

% PDB \cite{PDB}: model saliency using cascaded bidirectional LSTMs and using dilated convolutions 

% RVOS \cite{RVOS}: uses an RNN to select masks per frame and then to match them across frames 

% Learning Video Object Segmentation from Unlabeled Videos \cite{lu2020learning}

% AGNN \cite{AGNN}: Graph Neural Network, using attention, to model higher order relationships than simply 2-frame. 

% MATNet \cite{MAT}: Motion-Attentive Transition (MAT), is designed within a two-stream encoder, which
% transforms appearance features into motion-attentive representations at each convolutional stage.

\paragraph{Reinforcement Learning in Vision and Video.}  Reinforcement learning (RL) is conventionally used for tackling sequential decision making problems such as robot control or game-playing \cite{mnih2015dqn}. In recent years RL has increasingly been exploited in computer vision, where sequential decision problems also arise. For example: deciding which sequence of augmentation operators to apply in data augmentation pipelines \cite{cubuk2019autoAug}, sequence of image processing operators to apply in image restoration \cite{yu2018crafting}, or which sequence of words to predict in image captioning \cite{liu2017drlCaptioning}. Most related to our problem, RL has often been exploited in object tracking \cite{chen2018rlTracking,yun2017action}, where decisions about object locations and identities at time $t$ obviously affects the inference about these objects at time $t+1$. In the VOS problem, a related sequential decision-making problem arises as prior grouping decisions within a frame or at earlier frames affect the labelling decision for the subsequent regions. However, very few studies have attempted to apply RL to VOS. The main example thus far \cite{han2018reinforcement} essentially uses RL to solve a single object semi-supervised bounding-box tracking problem and then performs conventional segmentation within the tracked bounding box. In contrast, our framework performs multi-object VOS, directly optimised by RL. Please note that despite the similar name, \cite{goel2018unsupervised} is completely unrelated as it uses a fixed segmentation module to improve RL-game playing, rather than RL to improve segmentation. Besides enabling learning of sequential decision policies  RL also enables the optimization of non-differentiable rewards. This has been exploited in other tasks such as captioning  \cite{liu2017drlCaptioning} (to optimize language metrics) and tracking \cite{chen2018rlTracking,yun2017action} (to optimize IoU metrics). In this paper we exploit RL-based training to optimize such overlap-based metrics for video segmentation, rather than conventional pixel-wise cross-entropy.

\section{Method}

Our overall goal is to process a sequence of video frames $I_t$ with multiple moving objects, and generate a sequence of segmentation tensors $M_t$ that label each pixel with the consistent instance identity. So that each moving object in the video is tracked by a tube in $M_t$. 
Our framework generates object proposals and optical flow fields at every frame using off-the-shelf methods. These are fed to our selection network, that rejects stationary proposals and accepts moving proposals. Our assignment network then considers all proposals in one frame given the previous frame's segmentation, and makes a joint decision for how to group the current frame's proposals. Segmenting a video can thus be seen as a sequence of within-frame and across-frame grouping decisions. As a sequential decision problem, we train our model with reinforcement learning to optimise the total prediction-groundtruth overlap for the whole video. We explain each of these components in the following sections.

\subsection{Architecture}
\label{sec:architecture}

\begin{figure}[t]
\centering
\includegraphics[width=\linewidth]{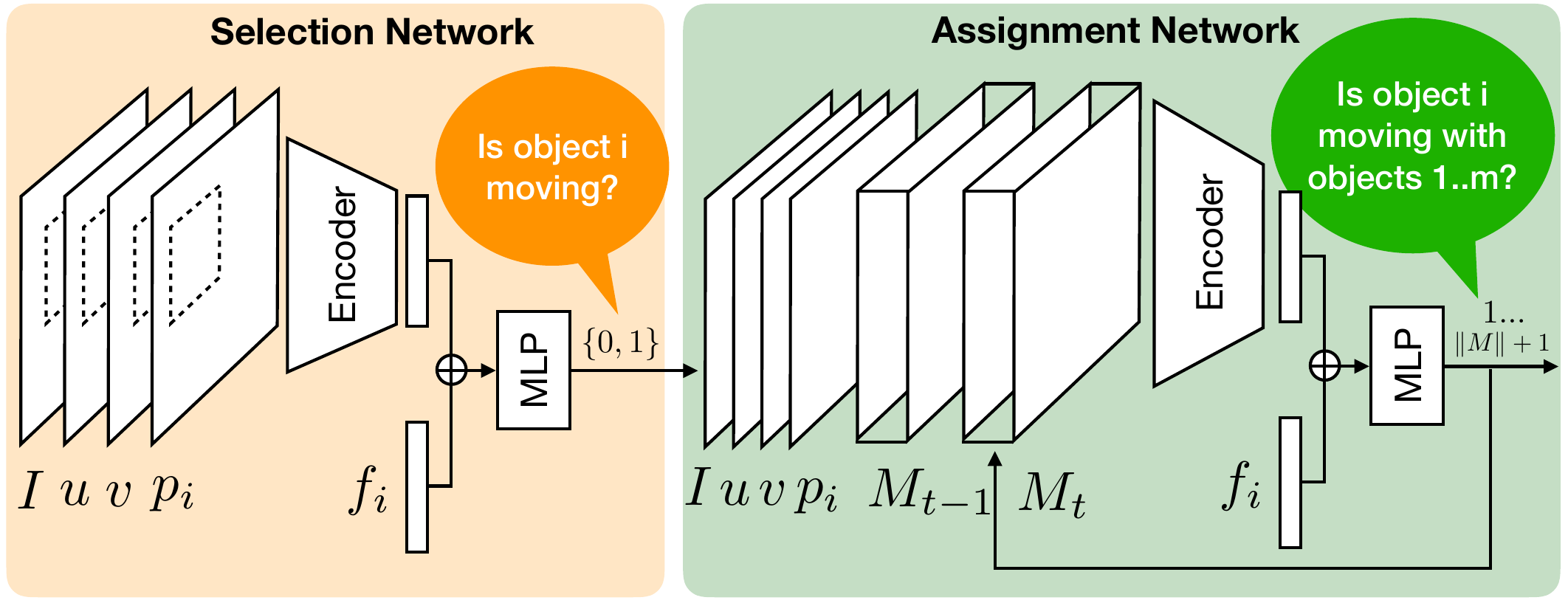}
\caption{Overview of the architecture. The selection network discriminates between moving and non-moving objects proposals ($p_i$). Given a moving object $i$, the assignment network assigns an ID based on coherent movement, appearance and shape, IDs assigned to objects in the previous frame $M_{t-1}$, and IDs assigned so far in the current frame $M_t$. }
\vspace{-0.1cm}
\label{fig:overview}
\end{figure}

The architecture is summarized in Fig.~\ref{fig:overview}. We process a sequence of images $I_t$. For each frame $t$ we generate an associated optical flow image $(u_t,v_t)$ along with a set of object proposals $\{p_i,f_i,b_i\}_t$, each described by a mask $p_i$, deep image feature $f_i$  and bounding box $b_i$. As output, we produce a segmentation tensor $M_t$ at every frame that labels each pixel with a persistent object ID or as background. For image size $w\times h$ and  max number of objects $K_{max}$ then $M_t\in\{1,0\}^{w\times h\times K_{max}}$. We can then project $M_t$ to a 2D segmentation in $\{0,\dots,K_{max}\}^{w\times h}$.

\paragraph{Selection Network. } We run an off-the-shelf category-agnostic proposal generator with low threshold to ensure all objects are considered with low false-positive rate. Our selection network then considers each proposal $i$ in turn, and accepts it as moving, or rejects it as stationary/background. It consists of a convolutional encoder module that inputs the depth concatenated RGB image, flow image, and proposal mask. The flow images are cropped to focus on the current region using its bounding box $b_i$. After processing by the selection encoder and pooling out spatial information, the appearance feature vector $f_i$ is fused before further processing by a MLP to produce a single binary output. This setup enables the selection network to detect moving foreground objects by performing spatial reasoning about shape (from the proposal), motion (from flow) and  appearance (RGB, deep feature), while avoiding distraction (due to cropping). 

%-- bounding box 
%-- encoder layers 
%-- MLP layers 

\paragraph{Assignment Network.}
The goal of the assignment network is to take the sequence of selected proposals $\mathbf{p}_t=\{p_{i,t}\}$ for each frame $t$ and group them into a consistent set of objects over time. If we denote the grouping label of proposal $i$ as $m_i$, then the goal at each frame is to make a joint decision about $p(M_t|\mathbf{p}_t,I_t,u_t,v_t,\mathbf{f}_t,M_{t-1})$ where $\mathbf{f}_t$ contains the appearance features for all selected proposals and $M_t=\{m_i\}_t$ denotes the labels of all proposals in the frame, or equivalently its segmentation. To make this joint inference, we use the tractable recurrent factorisation:
\begin{equation}\label{eq:assign}
    p(M_t|\mathbf{p}_t,I_t,u_t,v_t,\mathbf{f}_t,M_{t-1})=\prod_i p(m_{i,t}|M_{<i,t},{p}_{i,t},I_t,u_t,v_t,{f}_{i,t},M_{t-1})
\end{equation}

\noindent where $M_{<i,t}$ indicates the segmentation/grouping decisions in frame $t$ so far, prior to region $i$. That is, we consider each proposal $i$ in turn and assign it an object label based on its shape and appearance ($p_{i,t}, f_{i,t}$), the shared conditioning information ($I_t,u_t,v_t$), as well as the labels assigned to each proposal in the frame thus far $M_{<i,t}$ and labels in the prior frame $M_{t-1}$. 

To define the probability $p(m_{i,t}=k|p_{i,t},\dots)$ that a proposal $p_{i,t}$ is assigned to object $k\in \{1\dots K_{max}\}$, we take a softmax over logits $l^k_{i,t}$. Our assignment net predicts each logit $l^k_{i,t}$ by a Siamese relational network \cite{santoro2017relationNet,sung2018relationNet} that `compares' each proposal $i$ with putative grouping target $k$. Specifically, the assignment network contains a  CNN encoder module that inputs the depth concatenated images $(I_t,u_t,v_t,p_{i,t},M^k_{t-1},M^k_{<i,t})$
where $M_{<i,t}^k$ and $M^k_{t-1}$ denote taking the $k$-th object slice out of the corresponding tensors.  These are processed and spatially pooled, before fusing with the appearance feature $f_i$ by concatenation, and then fed into an MLP module that generates the logit $l^k_{i,t}$. 
%\doublecheck{Here $M_{t-1}$ is the 1-hot $w\times h\times K_{max}$ tensor describing the segmentation of the previous frame, and $M_t$ is the 1-hot $w\times h\times K_{max}$ tensor describing the segmentation \emph{so far} in the current frame.}
After the assignment network generates all the logits for one proposal we have defined $p(m_{i,t}|\dots)$ and we choose the max. Once an assignment decision is made, we update $M_{i,t}$ which is fed to the next iteration of Eq.~\ref{eq:assign}. The proposed masks are sorted by the confidence level of the proposal generator~\cite{maskrcnn}. In this way, we sequentially label each region in each frame, and each frame in the video. 
%\doublecheck{During training we sample this posterior, and during testing we maximise it. }

\subsection{Training}\label{sec:train}
\begin{figure}[t]
\centering
\includegraphics[width=0.55\textwidth]{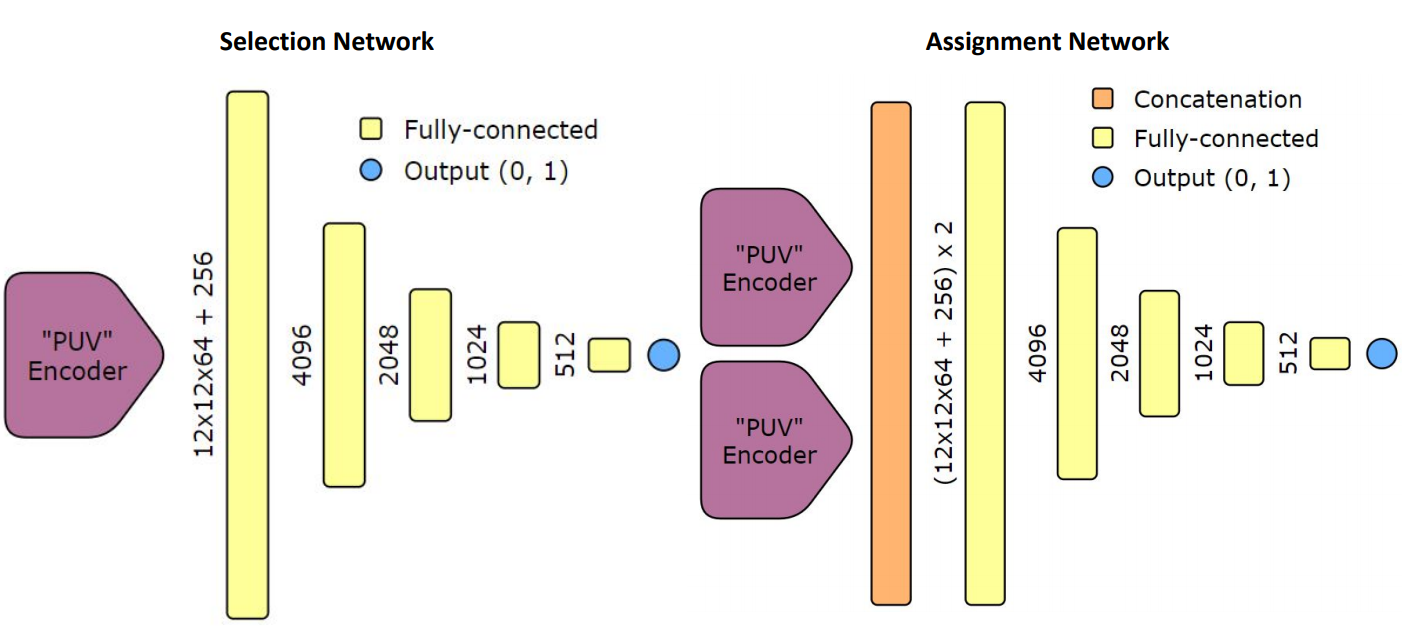}
\hspace{1cm}
\includegraphics[width=0.35\textwidth]{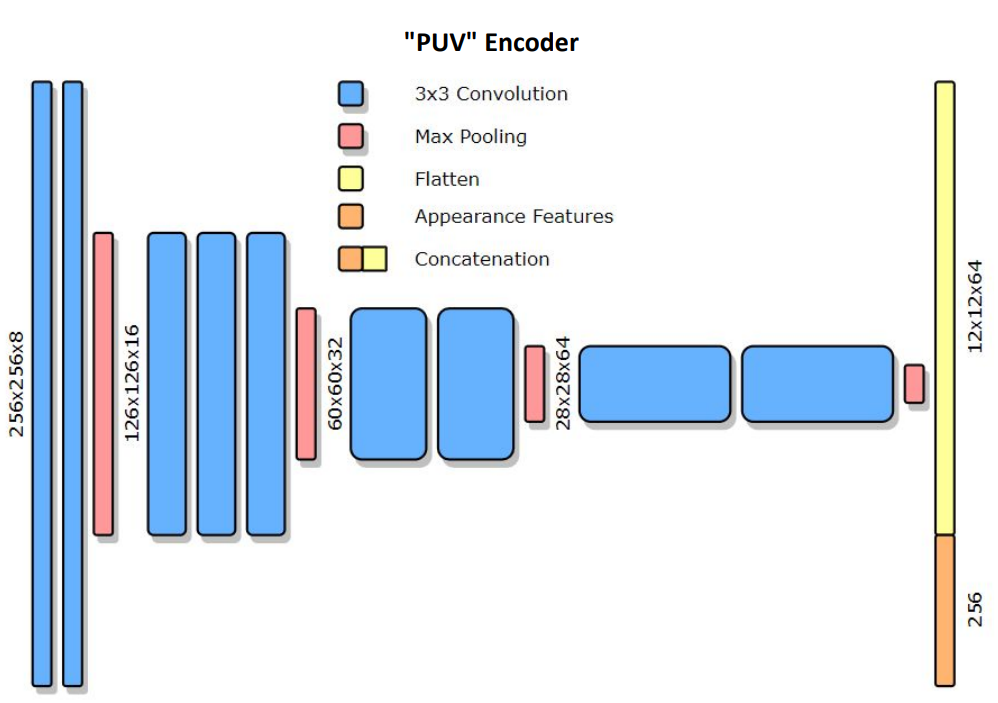}
\caption{Left: Selection and assignment networks. Right: The ``PUV" encoder network used as a module in the selection and assignment networks.``PUV" stands for the depth-concatenated mask proposals (P) with the optical flow (UV).}
\vspace{-0.1cm}
\label{fig:archi}
\end{figure}

We pre-train the selection and assignment networks with supervised learning as warm up, and then train the assignment network with reinforcement learning. Fig.~\ref{fig:archi} shows the architecture of the assignment and selection networks.

\paragraph{Ground Truth Generation.} 
In order to pre-train our network with supervised learning, we need ground truth labels for selection and assignment/grouping of the initial mask proposals. While the segmentation datasets provide ground truth assignments at the pixel level, this only helps us partially. For supervised learning at the proposal level, we need the optimal assignment given a set of mask proposals. Since this optimal assignment is costly to find, we use an approximation based on defining a greedy oracle with respect to the ground truth. We generate the selection network ground truth of whether objects are moving or not by iterating over the proposed masks and selecting those that overlap with any object instance in the ground truth more than 0.2. We generate the ground truth for assignments by iterating over the selected masks and matching those that improve the current IoU of any  the object instance. Note that this is not a perfect solution as different mask ordering (by confidence in our case) may lead to different solutions, and no ordering may be a global optima. We report the performance of this greedy oracle in the experiments. But note that this only applies to our supervised stage, our RL stage can potentially improve on this. 

\paragraph{Supervised Pre-Training.} 
Given the generated ``ground-truth", we train the selection net and assignment net by supervised learning with log-loss and cross-entropy respectively. After training, our selection network has 90\% accuracy (DAVIS 2017) on selecting true moving objects for subsequent consideration by the assignment network. We use a fixed 20 epoch pre-training.

\paragraph{Training with Reinforcement Learning.} RL algorithms are formalized by defining their state space, action space, transition function and reward. From this perspective, our state space consists of the set inputs to the assignment network including common (image, flow, etc) and specific to the current proposal (current mask, appearance, etc). Our action space is the set of possible grouping assignments for the current proposal. The transition function updates the state to provide the next proposal (within frame boundaries), and also next frame (across frame boundaries). It also updates the `grouping so far' $M_{<i,t}$ channel of the state with the results of the previous grouping action. Finally, we need to define the reward. The reward $R^v$ for video $v$ is the discounted sum of rewards per frame $R^v=\sum_t \gamma^t r^v_t$, where $r^v_t=\text{overlap}(M_t,GT_t)$, and $\gamma=0.99$. Overlap is quantified by $J\&F$-mean metric. We then use vanilla policy gradient \cite{williams1992reinforce}  training to optimize for the expected reward across all videos.

\paragraph{Implementation and Training Details.}
We choose as mask proposal generator the widely used MaskRCNN~\cite{maskrcnn}, without fine-tuning it. Note that this network is trained in the COCO dataset, where categories sometimes do not overlap with those found for example on the DAVIS 2017 dataset. Therefore, we use a low-confidence threshold of 0.05, to avoid discarding potential valid object instances. Even if the class is not correct because it was not in the training set, MaskRCNN tends to detect object-like regions. For the optical flow estimation we use the PWC-Net~\cite{pwcnet}. We enlarge the training set by doing some basic data augmentation, flipping all images horizontally. We train each of the selection and assignment networks for 30 epochs and use Adam optimizer with a learning rate of 1e-4. We use a batch size of 16 and reshape the flow and mask features to 256x256. For the RL part, we use a single agent that maximises the discounted reward. The encoder has 9 convolutional layers and 4 max-pooling layers and results in an output that corresponds to high level features. The MLP in the selection and assignment network consists of 5 FC layers each.

%-- MaskRCNN~\cite{maskrcnn} to generate proposals 
%-- PWC-Net~\cite{pwcnet} to generate flow fields
%-- DenseCRF to do postprocessing 
%-- Data Augmentation 
%-- 

\section{Experiments}

We show experimental results of the proposed \alg network. First, we test each of the proposed components in an ablation study, and observe that they all have a significant impact in the performance. In particular, training the network with RL is the component that improves most. We then compare the proposed method to current \sota methods, and observe that it outperforms all published previous work. 

%\subsection{Experimental Details}

\paragraph{Datasets and Settings.} We use three datasets for our experiments. We use the {\em DAVIS 2017}~\cite{davis2017} since it is the most widely used dataset for zero-shot, multi-object, instance segmentation benchmark. It contains 60 training and 30 validation sequences with 4209 and 1999 frames respectively. %There are 70.2 and 66.6 mean number of frames per sequence for training and validation. Total number of objects are 150 and 66 with an average of 2.4 and 2.2 objects per sequence for training and validation respectively. 
The performance metrics are region similarity or intersection over union $J$, and the boundary accuracy $F$. We use J\&F-Mean as the overlap metric for reward on all three datasets. We also use the very recent {\em YouTube-VOS}~\cite{vis}, which is the the largest video object segmentation dataset to date, and a very promising benchmark. While it does not include an official benchmark for the zero-shot segmentation problem, it can be easily adapted by simply not using the first frame as input. Previous zero-shot methods~\cite{RVOS} have also used that approach. The YouTube-VOS includes 3,471 and 474 videos in the training and validation set respectively. %\ls{is the metric also J and F?} % There are 65 unique seen object categories in the training and 91 unique object seen categories and 26 unseen categories in the validation set. In the zero-shot or unsupervised case,  the masks are not provided. Also, the objects to be segmented has to be discovered by the model.
Finally, we use the {\em FBMS}~\cite{fbms} dataset, which contains 59 video sequences, 29 are training and 30 for testing. %In comparison to the other two datasets we use, FBMS is a motion segmentation dataset.

\paragraph{Baselines.} For the ablation study we use the output from the object proposal generator (MaskRCNN~\cite{maskrcnn}) as a lower bound baseline. As an upper bound we use the generated ground truth described in Sec.~\ref{sec:architecture}. For the comparison to current \sota we simply use the top-performing methods from the DAVIS 2017 benchmark~\footnote{\url{https://davischallenge.org/davis2017/soa_compare.html}}, which include AGS~\cite{AGS}, RVOS~\cite{RVOS} and PDB~\cite{PDB}. It is worth noting that AGS is the top-performing of the three, but uses additional annotations of the saliency in DAVIS 2017. None of the other two methods or our own makes use of this additional labels. 

\subsection{Ablation Study}

We test the different components of the proposed \alg network and show the results in Table~\ref{table:ablation}. We start with the selection network (referred to as S in the table), which discards non-moving objects. The assignment is done by simply assigning a new object identity to each proposed mask. We observe that the selection network improves 4.4\% over the vanilla MaskRCNN results. While DAVIS 2017 does not penalize selecting non-moving objects, we attribute this improvement to reducing the number of possible mistakes at the assignment stage. We then show the results of adding a simple assignment network that takes in two proposals and predicts whether they belong to the same object. This assignment network is denoted A in the table. %\ls{shreyank, I forget if the recurrent connection was included or not...} 
Compared to the na{\"i}ve assignment, the simple assignment network improves significantly. We then add the more sophisticated assignment network from Fig.~\ref{fig:overview}, which includes the temporal component (T) $M_{t-1}$ and observe that this improves results further, due to object labels being more consistent over time. Finally, we train the assignment network using RL, as described in Sec.~\ref{sec:train}, and observe a 5.5\% increase in performance. Since the architecture is unchanged in this step, this is attributable to the switch form greedy (supervised) to non-myopic optimisation of assignment as well as optimisation of the target J\&F metric. This large jump shows the value of our contribution in terms of introducing RL to the VOS problem.

\begin{figure}[!h]
\centering
\includegraphics[width=\linewidth]{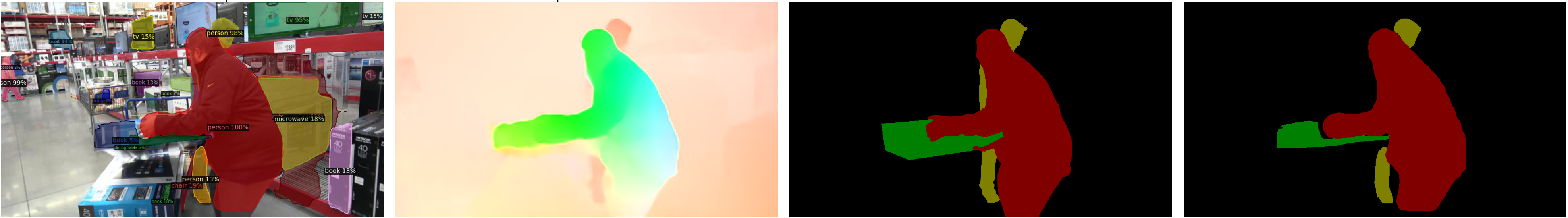}
\includegraphics[width=\linewidth]{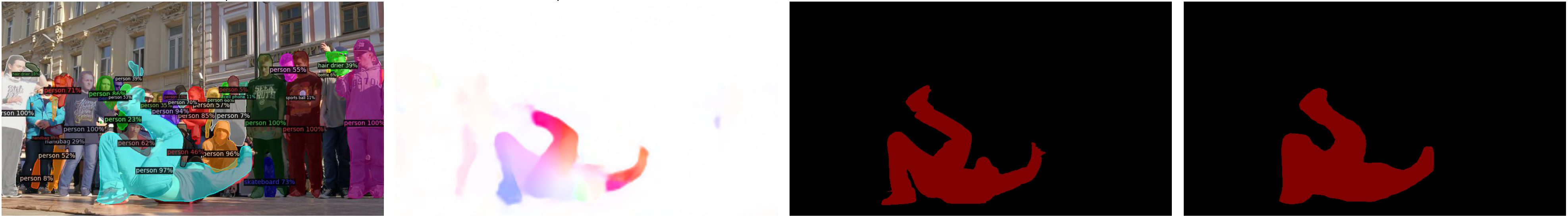}
\includegraphics[width=\linewidth]{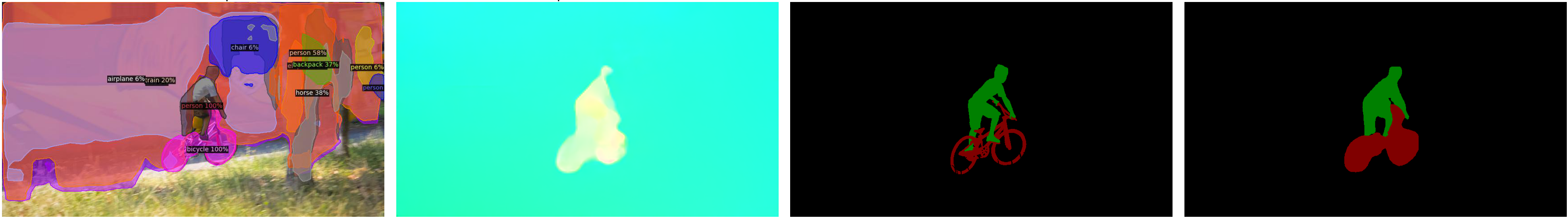}
\caption{Zero-shot segmentation results. From left to right: original image with MaskRCNN results super-imposed, optical flow estimation, ground truth, results from our method.}
\label{fig:samples_results}
\end{figure}

We also show some sample qualitative results in Fig.~\ref{fig:samples_results}. It is worth noting that the selection network correctly cleans up the initial proposals, for example in the case of crowded scenes like the breakdancing and the biking scene.
%It is also interesting that objects that are initially over segmented, like the box that is being passed in the first row, are successfully assigned to be the same object by the assignment network. 

\setlength{\tabcolsep}{4pt}
\begin{table}[t]
\begin{center}
\begin{tabular}{lccccccc} 
Method & J\&F-Mean & J-Mean & J-Recall & J-Decay & F-Mean & F-Recall & F-Decay \\ \hline
MRCNN & 38.9 & 37.0 & 42.3 & 0.3 & 40.9 & 43.0 & 2.5\\ \hline
S & 43.3 & 41.1 & 46.7 & {\bf -0.3}  & 45.5 & 47.8 & 5.0 \\
S+A & 49.6  & 49.3    & 52.7  & 4.3  & 49.9  &   51.8  &  2.5  \\
S+A+T & 52.9  & 52.4    & 55.1  & 4.1  & 53.4  &   54.6  &  {\bf 2.1}  \\
S+A+T+RL & {\bf 58.4} & {\bf 56.6} & {\bf 63.4} & 7.7 & {\bf 60.2} & {\bf 63.1} &  7.9 \\ \hline
%S+A+T+RL+P & & & & & & & \\ 
Oracle & 64.1 & 62.9 & 72.0 & -2.0 & 65.3 & 73.1 & 1.7\\ \hline
\end{tabular}
\end{center}
\caption{{\bf S}: Selection network, {\bf A}: Assignment Network, {\bf T}: Temporal information, {\bf RL}: Optimized with Reinforcement Learning. Results of different components of the proposed network on DAVIS 2017 validation set.  }
\label{table:ablation}
\end{table}

\begin{figure}[!h]
\centering
\includegraphics[width=\linewidth]{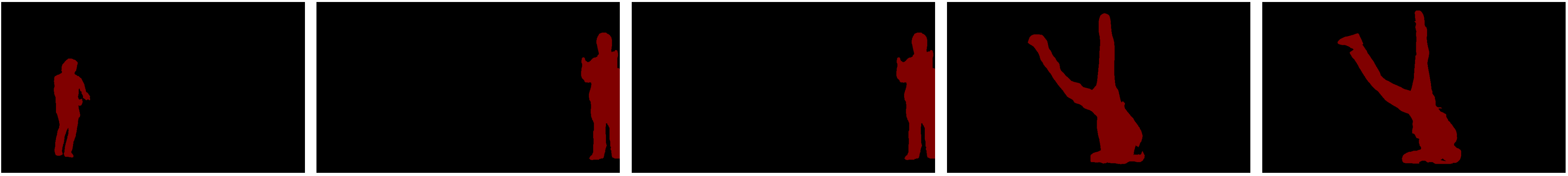}
\includegraphics[width=\linewidth]{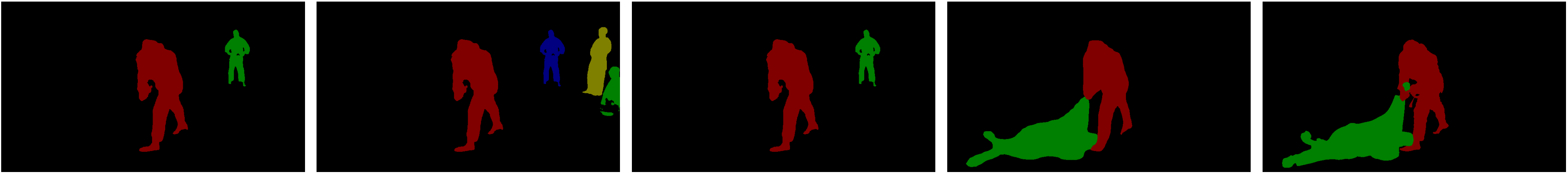}
\includegraphics[width=\linewidth]{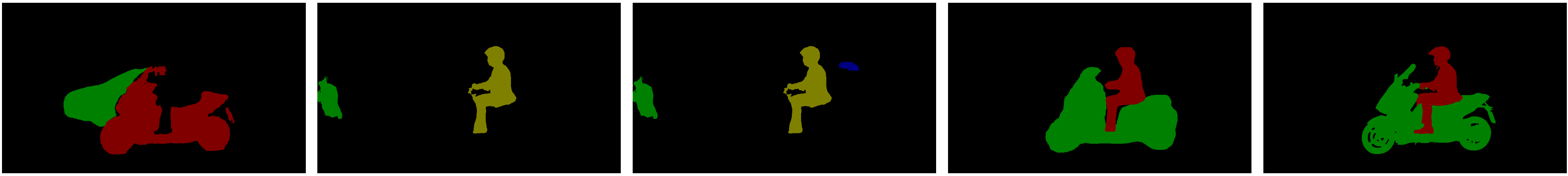}
\includegraphics[width=\linewidth]{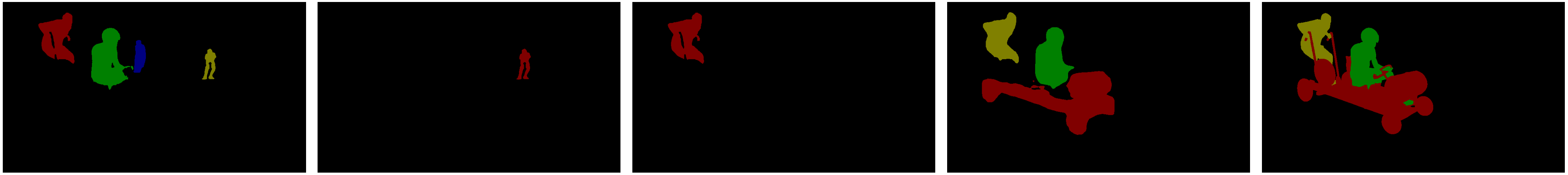}
\includegraphics[width=\linewidth]{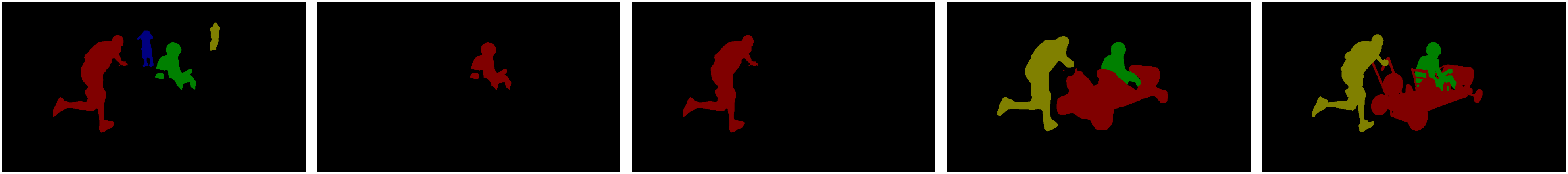}
\caption{Qualitative comparison to \sota on DAVIS 2017. From left to right: AGNN, PDB, AGS, \alg, ground truth. }
\label{fig:samples_sota}
\end{figure}

\subsection{Comparison to State-of-the-art}

We now compare the full proposed \alg method to the current \sota. The numerical results are shown in Table~\ref{table:davis2017} for DAVIS2017, Table~\ref{table:FBMS} for FBMS and Table~\ref{table:youtubeVOS} for Youtube-VOS respectively. The results show that \alg consistently outperforms alternatives across different datasets, although they come from different sources and thus expose different challenges. For example, the criteria for different datasets are slightly different in each dataset; objects that move together are grouped together in FBMS, while objects that are semantically different are distinct in the others -- creating differences for person plus horse type examples. Meanwhile Youtube-VOS introduces unseen object classes between training and testing. Qualitative results comparing \alg to \sota alternatives on the DAVIS dataset are shown in Fig.~\ref{fig:samples_sota}. 

\setlength{\tabcolsep}{4pt}
\begin{table}[h]
\begin{center}
\begin{tabular}{lccccccccc} 
Method & \specialcell{Add.\\Data} & \specialcell{Post \\Process.} & \specialcell{J\&F\\Mean} & \specialcell{J\\Mean} & \specialcell{J\\Recall} & \specialcell{J\\Decay} & \specialcell{F\\Mean} & \specialcell{F\\Recall} & \specialcell{F\\Decay} \\ \hline
PDB~\cite{PDB} & \xmark & \cmark & 55.1 & 53.2 & 58.9 & 4.9 & 57.0 & 60.2 & 6.8  \\
RVOS~\cite{RVOS} & \xmark & \xmark & 41.2 & 36.8 & 40.2 & 0.5 & 45.7 & 46.4 & 1.7  \\
\alg & \xmark & \xmark & {\bf 58.4} & {\bf 56.6} & {\bf 63.4} & 7.7 & {\bf 60.2} & {\bf 63.1} &  7.9  \\ \hline
AGS~\cite{AGS} & \cmark & \cmark & 57.5 & 55.5 & 61.6 & 7.0 & 59.5 & 62.8 & 9.0 
\\ \hline
\end{tabular}
\end{center}
\caption{Comparison to other \sota on the DAVIS 2017 validation set.  }
\label{table:davis2017}
\end{table}

\setlength{\tabcolsep}{5pt}
\begin{table}[h]
\begin{center}
\begin{tabular}{lccccc} 
Method & MAT~\cite{MAT} & MBN~\cite{li2018unsupervised} & PDB~\cite{PDB} & IET~\cite{li2018instance} & \alg \\ \hline
J-mean & 76.1 & 73.9 & 74.0 & 71.9 & \textbf{77.6} \\
F-score & - & 83.2 & \textbf{84.9} & 82.8 & 84.4  \\ \hline
% Method & J-mean & F-score\\ \hline
% MAT~\cite{MAT} & 76.1 & - \\
% MBN~\cite{li2018unsupervised} & 73.9 & 83.2 \\
% PDB~\cite{PDB} & 74.0 & \textbf{84.9} \\
% IET~\cite{li2018instance} & 71.9 & 82.8 \\
% \alg & \textbf{77.6} & 84.4 \\ \hline
\end{tabular}
\end{center}
\caption{Results on FBMS dataset.}
\label{table:FBMS}
\end{table}

\setlength{\tabcolsep}{5pt}
\begin{table}[h]
\begin{center}
\begin{tabular}{lcccc} 
Method & J-seen & J-unseen & F-seen & F-unseen\\ \hline
RVOS~\cite{RVOS} & 44.7 & 21.2 & 45.0 & 23.9 \\
\alg & \textbf{53.8} & \textbf{34.4} & \textbf{51.6} & \textbf{35.8} \\ \hline
\end{tabular}
\end{center}
\caption{Results on Youtube-VOS dataset.}
\label{table:youtubeVOS}
\end{table}

% \setlength{\tabcolsep}{5pt}
% \begin{table}[t]
% \begin{center}
% \begin{tabular}{lcc} 
% Method & J-mean & F-score\\ \hline
% MAT~\cite{MAT} & 76.1 & - \\
% MBN~\cite{li2018unsupervised} & 73.9 & 83.2 \\
% PDB~\cite{PDB} & 74.0 & \textbf{84.9} \\
% IET~\cite{li2018instance} & 71.9 & 82.8 \\
% \alg & \textbf{77.6} & 84.4 \\ \hline
% \end{tabular}
% \end{center}
% \caption{Results on FBMS dataset.}
% \label{table:FBMS}
% \end{table}

\section{Conclusion}

We considered the zero-shot video object segmentation task as a selection and grouping problem over generic object proposals, and discussed how training such a model with RL enables better global decision-making and direct optimization of overlap metrics. We showed that our grouping architecture surpasses previous approaches when trained with RL. We believe that this is the first demonstration of RL in VOS and hope that it leads others to leverage this technique for VOS, VIS and related tasks in future.

%\pagebreak

\bibliography{egbib}

\begin{thebibliography}{32}
\providecommand{\natexlab}[1]{#1}
\providecommand{\url}[1]{\texttt{#1}}
\expandafter\ifx\csname urlstyle\endcsname\relax
  \providecommand{\doi}[1]{doi: #1}\else
  \providecommand{\doi}{doi: \begingroup \urlstyle{rm}\Url}\fi

\bibitem[Bengio et~al.(2015)Bengio, Vinyals, Jaitly, and
  Shazeer]{bengio2015scheduled}
Samy Bengio, Oriol Vinyals, Navdeep Jaitly, and Noam Shazeer.
\newblock Scheduled sampling for sequence prediction with recurrent neural
  networks.
\newblock In \emph{NIPS}, 2015.

\bibitem[Caelles et~al.(2019)Caelles, Pont-Tuset, Perazzi, Montes, Maninis, and
  {Van Gool}]{davis2017}
Sergi Caelles, Jordi Pont-Tuset, Federico Perazzi, Alberto Montes,
  Kevis-Kokitsi Maninis, and Luc {Van Gool}.
\newblock The 2019 davis challenge on vos: Unsupervised multi-object
  segmentation.
\newblock \emph{arXiv:1905.00737}, 2019.

\bibitem[Chen et~al.(2018)Chen, Wang, Li, Wang, and Lu]{chen2018rlTracking}
Boyu Chen, Dong Wang, Peixia Li, Shuang Wang, and Huchuan Lu.
\newblock Real-time 'actor-critic' tracking.
\newblock In \emph{ECCV}, 2018.

\bibitem[Cubuk et~al.(2019)Cubuk, Zoph, Man{\'{e}}, Vasudevan, and
  Le]{cubuk2019autoAug}
Ekin~Dogus Cubuk, Barret Zoph, Dandelion Man{\'{e}}, Vijay Vasudevan, and
  Quoc~V. Le.
\newblock Autoaugment: Learning augmentation policies from data.
\newblock \emph{CVPR}, 2019.

\bibitem[Goel et~al.(2018)Goel, Weng, and Poupart]{goel2018unsupervised}
Vikash Goel, Jameson Weng, and Pascal Poupart.
\newblock Unsupervised video object segmentation for deep reinforcement
  learning.
\newblock In \emph{NeurIPS}, 2018.

\bibitem[Han et~al.(2018)Han, Yang, Zhang, Chang, and
  Liang]{han2018reinforcement}
Junwei Han, Le~Yang, Dingwen Zhang, Xiaojun Chang, and Xiaodan Liang.
\newblock Reinforcement cutting-agent learning for video object segmentation.
\newblock In \emph{CVPR}, 2018.

\bibitem[{He} et~al.(2017){He}, {Gkioxari}, {Dollár}, and
  {Girshick}]{maskrcnn}
K.~{He}, G.~{Gkioxari}, P.~{Dollár}, and R.~{Girshick}.
\newblock Mask {R-CNN}.
\newblock In \emph{ICCV}, 2017.

\bibitem[Jain et~al.(2017)Jain, Xiong, and Grauman]{fseg}
Suyog Jain, Bo~Xiong, and Kristen Grauman.
\newblock Fusionseg: Learning to combine motion and appearance for fully
  automatic segmention of generic objects in videos.
\newblock In \emph{CVPR}, 2017.

\bibitem[Li et~al.(2018{\natexlab{a}})Li, Seybold, Vorobyov, Fathi, Huang, and
  Jay~Kuo]{li2018instance}
Siyang Li, Bryan Seybold, Alexey Vorobyov, Alireza Fathi, Qin Huang, and C-C
  Jay~Kuo.
\newblock Instance embedding transfer to unsupervised video object
  segmentation.
\newblock In \emph{CVPR}, 2018{\natexlab{a}}.

\bibitem[Li et~al.(2018{\natexlab{b}})Li, Seybold, Vorobyov, Lei, and
  Jay~Kuo]{li2018unsupervised}
Siyang Li, Bryan Seybold, Alexey Vorobyov, Xuejing Lei, and C-C Jay~Kuo.
\newblock Unsupervised video object segmentation with motion-based bilateral
  networks.
\newblock In \emph{ECCV}, 2018{\natexlab{b}}.

\bibitem[Li and Change~Loy(2018)]{li2018video}
Xiaoxiao Li and Chen Change~Loy.
\newblock Video object segmentation with joint re-identification and
  attention-aware mask propagation.
\newblock In \emph{ECCV}, 2018.

\bibitem[Liu et~al.(2017)Liu, Zhu, Ye, Guadarrama, and
  Murphy]{liu2017drlCaptioning}
Siqi Liu, Zhenhai Zhu, Ning Ye, Sergio Guadarrama, and Kevin Murphy.
\newblock Improved image captioning via policy gradient optimization of spider.
\newblock In \emph{ICCV}, 2017.

\bibitem[Maninis et~al.(2018)Maninis, Caelles, Chen, Pont-Tuset, Leal-Taix\'e,
  Cremers, and {Van Gool}]{osvos}
Kevis-Kokitsi Maninis, Sergi Caelles, Yuhua Chen, Jordi Pont-Tuset, Laura
  Leal-Taix\'e, Daniel Cremers, and Luc {Van Gool}.
\newblock Video object segmentation without temporal information.
\newblock \emph{IEEE Transactions on Pattern Analysis and Machine Intelligence
  (TPAMI)}, 2018.

\bibitem[Mnih et~al.(2015)Mnih, Kavukcuoglu, Silver, Rusu, Veness, Bellemare,
  Graves, Riedmiller, Fidjeland, Ostrovski, Petersen, Beattie, Sadik,
  Antonoglou, King, Kumaran, Wierstra, Legg, and Hassabis]{mnih2015dqn}
Volodymyr Mnih, Koray Kavukcuoglu, David Silver, Andrei~A. Rusu, Joel Veness,
  Marc~G. Bellemare, Alex Graves, Martin Riedmiller, Andreas~K. Fidjeland,
  Georg Ostrovski, Stig Petersen, Charles Beattie, Amir Sadik, Ioannis
  Antonoglou, Helen King, Dharshan Kumaran, Daan Wierstra, Shane Legg, and
  Demis Hassabis.
\newblock Human-level control through deep reinforcement learning.
\newblock \emph{Nature}, 518\penalty0 (7540):\penalty0 529--533, 02 2015.

\bibitem[Ranzato et~al.(2016)Ranzato, Chopra, Auli, and
  Zaremba]{renzato2016sequenceTrainingRNN}
Marc'Aurelio Ranzato, Sumit Chopra, Michael Auli, and Wojciech Zaremba.
\newblock Sequence level training with recurrent neural networks.
\newblock In \emph{ICLR}, 2016.

\bibitem[Santoro et~al.(2017)Santoro, Raposo, Barrett, Malinowski, Pascanu,
  Battaglia, and Lillicrap]{santoro2017relationNet}
Adam Santoro, David Raposo, David G.~T. Barrett, Mateusz Malinowski, Razvan
  Pascanu, Peter~W. Battaglia, and Timothy~P. Lillicrap.
\newblock A simple neural network module for relational reasoning.
\newblock In \emph{NIPS}, 2017.

\bibitem[Song et~al.(2018)Song, Wang, Zhao, Shen, and Lam]{PDB}
Hongmei Song, Wenguan Wang, Sanyuan Zhao, Jianbing Shen, and Kin-Man Lam.
\newblock Pyramid dilated deeper convlstm for video salient object detection.
\newblock In \emph{ECCV}, 2018.

\bibitem[Sun et~al.(2018)Sun, Yang, Liu, and Kautz]{pwcnet}
Deqing Sun, Xiaodong Yang, Ming-Yu Liu, and Jan Kautz.
\newblock {PWC-Net}: {CNNs} for optical flow using pyramid, warping, and cost
  volume.
\newblock In \emph{CVPR}, 2018.

\bibitem[Sung et~al.(2018)Sung, Yang, Zhang, Xiang, Torr, and
  Hospedales]{sung2018relationNet}
Flood Sung, Yongxin Yang, Li~Zhang, Tao Xiang, Philip H.~S. Torr, and
  Timothy~M. Hospedales.
\newblock Learning to compare: Relation network for few-shot learning.
\newblock In \emph{CVPR}, 2018.

\bibitem[T.Brox and J.Malik(2010)]{fbms}
T.Brox and J.Malik.
\newblock Object segmentation by long term analysis of point trajectories.
\newblock In \emph{ECCV}, 2010.

\bibitem[Tokmakov et~al.(2017)Tokmakov, Alahari, and Schmid]{LVO}
P.~Tokmakov, K.~Alahari, and C.~Schmid.
\newblock Learning video object segmentation with visual memory.
\newblock In \emph{ICCV}, 2017.

\bibitem[Ventura et~al.(2019)Ventura, Bellver, Girbau, Salvador, Marques, and
  Giro-i Nieto]{RVOS}
Carles Ventura, Miriam Bellver, Andreu Girbau, Amaia Salvador, Ferran Marques,
  and Xavier Giro-i Nieto.
\newblock {RVOS}: End-to-end recurrent network for video object segmentation.
\newblock In \emph{CVPR}, 2019.

\bibitem[Wang et~al.(2019{\natexlab{a}})Wang, Lu, Shen, Crandall, and
  Shao]{AGNN}
Wenguan Wang, Xiankai Lu, Jianbing Shen, David~J Crandall, and Ling Shao.
\newblock Zero-shot video object segmentation via attentive graph neural
  networks.
\newblock In \emph{ICCV}, 2019{\natexlab{a}}.

\bibitem[Wang et~al.(2019{\natexlab{b}})Wang, Song, Zhao, Shen, Zhao, Hoi, and
  Ling]{AGS}
Wenguan Wang, Hongmei Song, Shuyang Zhao, Jianbing Shen, Sanyuan Zhao,
  Steven~CH Hoi, and Haibin Ling.
\newblock Learning unsupervised video object segmentation through visual
  attention.
\newblock In \emph{CVPR}, 2019{\natexlab{b}}.

\bibitem[Williams(1992)]{williams1992reinforce}
Ronald~J Williams.
\newblock Simple statistical gradient-following algorithms for connectionist
  reinforcement learning.
\newblock \emph{Machine Learning}, 1992.

\bibitem[Xu et~al.(2019)Xu, Liu, Bao, Liu, and Zhou]{xu2019mhp}
Shuangjie Xu, Daizong Liu, Linchao Bao, Wei Liu, and Pan Zhou.
\newblock {MHP-VOS}: Multiple hypotheses propagation for video object
  segmentation.
\newblock In \emph{CVPR}, 2019.

\bibitem[Yang et~al.(2019{\natexlab{a}})Yang, Fan, and Xu]{vis}
Linjie Yang, Yuchen Fan, and Ning Xu.
\newblock Video instance segmentation.
\newblock In \emph{ICCV}, 2019{\natexlab{a}}.

\bibitem[Yang et~al.(2019{\natexlab{b}})Yang, Wang, Bertinetto, Bai, Hu, and
  Torr]{andiff}
Zhao Yang, Qiang Wang, Luca Bertinetto, Song Bai, Weiming Hu, and Philip~H.S.
  Torr.
\newblock Anchor diffusion for unsupervised video object segmentation.
\newblock In \emph{ICCV}, 2019{\natexlab{b}}.

\bibitem[Yu et~al.(2018)Yu, Dong, Lin, and Change~Loy]{yu2018crafting}
Ke~Yu, Chao Dong, Liang Lin, and Chen Change~Loy.
\newblock Crafting a toolchain for image restoration by deep reinforcement
  learning.
\newblock In \emph{CVPR}, 2018.

\bibitem[Yun et~al.(2017)Yun, Choi, Yoo, Yun, and Young~Choi]{yun2017action}
Sangdoo Yun, Jongwon Choi, Youngjoon Yoo, Kimin Yun, and Jin Young~Choi.
\newblock Action-decision networks for visual tracking with deep reinforcement
  learning.
\newblock In \emph{CVPR}, 2017.

\bibitem[Zeng et~al.(2019)Zeng, Liao, Gu, Xiong, Fidler, and
  Urtasun]{zeng2019dmm}
Xiaohui Zeng, Renjie Liao, Li~Gu, Yuwen Xiong, Sanja Fidler, and Raquel
  Urtasun.
\newblock {DMM-Net}: Differentiable mask-matching network for video object
  segmentation.
\newblock In \emph{CVPR}, 2019.

\bibitem[Zhou et~al.(2020)Zhou, Wang, Zhou, Yao, Li, and Shao]{MAT}
Tianfei Zhou, Shunzhou Wang, Yi~Zhou, Yazhou Yao, Jianwu Li, and Ling Shao.
\newblock Motion-attentive transition for zero-shot video object segmentation.
\newblock In \emph{AAAI}, 2020.

\end{thebibliography}
\end{document}